\documentclass[letterpaper]{article} % DO NOT CHANGE THIS
\usepackage{aaai25}  % DO NOT CHANGE THIS
\usepackage{times}  % DO NOT CHANGE THIS
\usepackage{helvet}  % DO NOT CHANGE THIS
\usepackage{courier}  % DO NOT CHANGE THIS
\usepackage[hyphens]{url}  % DO NOT CHANGE THIS
\usepackage{graphicx} % DO NOT CHANGE THIS
\usepackage{natbib}  % DO NOT CHANGE THIS AND DO NOT ADD ANY OPTIONS TO IT
\usepackage{caption} % DO NOT CHANGE THIS AND DO NOT ADD ANY OPTIONS TO IT
\usepackage{algorithm}
\usepackage{algorithmic}

\usepackage{adjustbox}
\usepackage{amssymb}
\usepackage{booktabs}
\usepackage{multirow,multicol}
\usepackage{tablefootnote}
\usepackage{subcaption}
\usepackage{tabularx}
\usepackage{verbatimbox}
\usepackage[noabbrev,capitalise]{cleveref}

\usepackage{xcolor}
\newcommand{\answerYes}[1]{\textcolor{blue}{#1}} 
\newcommand{\answerNo}[1]{\textcolor{teal}{#1}} 
\newcommand{\answerNA}[1]{\textcolor{gray}{#1}}

\newcommand{\mc}{\multicolumn}
\newcommand{\mr}{\multirow}

\newcommand{\tc}[1]{\mc{1}{c}{#1}}  % table center a single cell
\newcommand{\ck}{\checkmark}

\usepackage{newfloat}
\usepackage{listings}
\DeclareCaptionStyle{ruled}{labelfont=normalfont,labelsep=colon,strut=off} % DO NOT CHANGE THIS
\floatstyle{ruled}
\newfloat{listing}{tb}{lst}{}
\floatname{listing}{Listing}
\nocopyright % -- Your paper will not be published if you use this command
\title{Exploring Vision Language Models for\\
       Multimodal and Multilingual Stance Detection}
\author {
    Jake Vasilakes\textsuperscript{\rm 1},
    Carolina Scarton\textsuperscript{\rm 1},
    Zhixue Zhao\textsuperscript{\rm 1}
}
\affiliations {
    \textsuperscript{\rm 1}Department of Computer Science, University of Sheffield, UK\\
    \{j.vasilakes, c.scarton, zhixue.zhao\}@sheffield.ac.uk
}

\begin{document}

\maketitle

\begin{abstract}
Social media’s global reach amplifies the spread of information, highlighting the need for robust Natural Language Processing tasks, like stance detection, across languages and modalities. Prior research predominantly focuses on text-only inputs, leaving multimodal scenarios, such as those involving both images and text, relatively underexplored. Meanwhile, the prevalence of multimodal posts has increased significantly in recent years. Although state-of-the-art Vision-Language Models (VLMs) show promise, their performance on multimodal and multilingual stance detection tasks remains largely unexamined. This paper evaluates state-of-the-art VLMs on a newly extended dataset covering seven languages and multimodal inputs, investigating their use of visual cues, language-specific performance, and cross-modality interactions. Our results show that VLMs generally rely more on text than images for stance detection and this trend persists across languages. Additionally, VLMs rely significantly more on text contained within the images than other visual content. Regarding multilinguality, the models studied tend to generate consistent predictions across languages whether they are explicitly multilingual or not, although there are outliers that are incongruous with macro F1, language support, and model size.\footnote{We will make our code and data publicly available upon publication.}
\end{abstract}


\section{Introduction}

The ubiquity of social media has led to an increased reliance on these platforms for news and science communication, often surpassing traditional, slower-to-publish sources such as print news media and peer-reviewed scientific journals \cite{Matsa_2015,Gürer_Hubbard_Bohon_2023}. A side-effect of this is an increase in the speed of information spread, including rumors and fake news \cite{Ceylan_Anderson_Wood_2023}. Understanding the spread of information on social media requires knowing the viewpoints users take with respect to claims, topics, or entities. This is the purview of stance detection, a key task within Natural Language Processing (NLP), which aims to automatically classify an author's viewpoint with respect to a specific target.

While stance detection has been studied extensively on English text, efforts to extend it to modalities beyond text and languages beyond English are relatively sparse (cf. \citet{Küçük_Can_2020} sections 9.1 and 9.2). Additionally, the studies that do go beyond tend to investigate only one or the other, and in the case of other languages tend to focus on one or a few languages at a time. One potential reason is that obtaining high quality data requires significant time and resources. Still, extending stance detection systems to other modalities and languages is crucial, as rumors on social media often spread across languages \cite{Singh_Bontcheva_Scarton_2024}, and posts often make use of images or videos to reinforce their message. For example, a user might comment on a news segment, use an image as supporting evidence for a scientific claim, or post a meme that expresses their view of a political figure.

Recent development of Vision Language Models (VLMs), available within easy-to-use tools such as ChatGPT\footnote{\url{https://openai.com/index/dall-e-3}} and Huggingface\footnote{\url{https://huggingface.co}}, provides powerful and accessible means to perform stance detection using both text and images. Furthermore, the Large Language Models (LLMs) that form the backbone of VLMs are often capable of processing and analyzing text in a variety of languages. However, to our knowledge, there has been no study investigating the ability of VLMs to perform stance detection using both text and image modalities, nor has there been any examination of their performance on this task across languages.

Given this research gap, this paper investigates the performance of state-of-the-art VLMs for stance detection at the intersection of modalities beyond text and languages beyond English. Our specific contributions are threefold.

\begin{itemize}
    \item \textbf{Image Use in Multimodality}: an analysis of the extent to which VLMs can effectively use information from images when performing stance detection.
    \item \textbf{Multilinguality}: an investigation into the performance of VLMs on stance detection across languages.
    \item \textbf{Joint Multimodality and Multilinguality}: an exploration of the interaction between text and images across languages in VLMs for stance detection.
\end{itemize}

Our experiments make use of a recently developed multimodal stance detection dataset in English, which we extend to cover six additional languages \cite{Liang_et_al_2024}.

\section{Background}

We begin with an overview of the stance detection task, highlighting previous efforts on multimodal and multilingual approaches, including a discussion of currently available datasets. We then provide a review of the general architecture of VLMs, as well as previous research that attempts to use them for stance detection.

\subsection{Stance Detection}

Stance detection is the task of automatically determining the viewpoint, position, or sentiment of an author regarding a target. This target could be a topic (e.g., a news event or scientific theory), an entity (e.g., a political figure), or even the viewpoint of another author (e.g., another tweet expressing a viewpoint about a political figure) \cite{Küçük_Can_2020}. Generally, stance detection is formulated as a paired text classification problem where the input is a stance text and a target text (e.g., a tweet and the name of a political figure) and the output is one of \textit{Favour, Against}, or \textit{Neutral}.

Stance detection is fundamental to wrangling the spread of rumors online. Interest in this task has therefore grown alongside the increased popularity of social media, which has both lowered the barrier to content publishing and increased the speed at which content can be disseminated \cite{ALDayel_Magdy_2021}. Early efforts focused on supervised classification with SVMs \cite{elfardy2016cu,mohammad2017stance}, naive Bayes \cite{walker2012your}, or neural networks \cite{siddiqua2019tweet,li2019multi}, which led to models based on pretrained transformers \cite{fajcik-etal-2019-fit,kawintiranon-singh-2021-knowledge,Khandelwal_2021}.
The recent development of LLMs has introduced the possibility of 0-shot stance detection, although research so far has evaluated mostly GPT models \cite{lan2024stance,liyanage2023gpt,suppa-etal-2024-bryndza}.  The exception is \citet{cruickshank2023use} who performed evaluations with both 0-shot and fine-tuned LLMs, finding that fine-tuning does not necessarily increase performance.

Research on multimodal and multilingual stance detection is more sparse. There are only a few studies using images alongside text \cite{Hu_Liu_Wang_Zhang_Lin_2023,Wang_Zuo_Peng_Plank_2024,Niu_et_al_2024}, including \citet{Liang_et_al_2024} which introduced the dataset used in this work. Stance detection on languages besides English has generally focused on one or a few languages at a time. These include \citet{Zotova_Agerri_Nuñez_Rigau_2020} (Catalan and Spanish), \citet{Vamvas_Sennrich_2020} (German, French, Italian), \citet{Alhindi_et_al_2021} (Arabic), and \citet{Zheng_Baheti_Naous_Xu_Ritter_2022} (Hindi, Arabic). To our knowledge the only exception is the COFe dataset, which contains comments from a debate platform covering 26 languages \cite{barriere-etal-2022-cofe}.

\subsection{Vision Language Models}

Vision Language Models (VLMs) are an extension of LLMs to images and videos. Generally, they are comprised of a vision model and an LLM. The vision model encodes the image input, which is then projected using a neural network into a latent space consistent with the text embeddings of the LLM. These ``image tokens'' are then concatenated with the text tokens and input to the LLM which generates text output. An in-depth review of VLMs can be found in \citet{Yin_et_al_2024}.

To the best of our knowledge, the only existing evaluation of VLMs on multimodal stance detection is given in \citet{Liang_et_al_2024}, who evaluate Qwen-VL and GPT4-Vision. While they compare these to a variety of other text-only, vision-only, and multimodal models, these belong to different model families (e.g., BERT \cite{Devlin_Chang_Lee_Toutanova_2019}, ViT \cite{Dosovitskiy_et_al_2021}, and CLIP \cite{Radford_et_al_2021}). It is thus not possible to directly compare their results to gain insight into models' use of the text and image modalities, as we aim to do here.

\section{Evaluation Methodology}

This section describes the dataset used in our experiments, how we chose which models to evaluate, and an overview of each of our three experiments.

\subsection{Dataset}

Despite the importance of treating text and images together when performing stance detection on social media data, there is a lack of datasets that include both modalities, not to mention both modalities covering languages other than English. We therefore use the dataset introduced by \citet{Liang_et_al_2024}, which contains tweet-image pairs in English collected from X covering 5 news topics. A summary of the dataset is given in \cref{tab:dataset}.

\begin{table}[]
    \centering
    \small
    \begin{tabularx}{\linewidth}{X c rrrr}
        \toprule
         \textbf{Topic}                               & \textbf{Target}          & \textbf{\#}  & \textbf{F}   &  \textbf{A}   &  \textbf{N}    \\
        \midrule
        COVID-19                             & Cloroquinine    & 141 & 35.5\% & 33.3\% & 31.2\%  \\
        \midrule
        \mr{2}{2cm}{Russia- Ukraine}          & Russia          & 111 & 2.7\% & 67.6\% & 29.7\% \\
                                             & Ukraine         & 108 & 60.2\% & 3.7\% & 36.1\% \\
        \midrule
        \mr{2}{2cm}{2020 US Election}        & Trump           & 170 & 78.9\% & 14.1\% & 7.1\%  \\ 
                                             & Biden           & 128 & 50.8\% & 37.5\% & 11.7\% \\
        \midrule
        \mr{2}{2cm}{Taiwan Question}         & China           & 140 & 24.3\% & 62.1\% & 13.5\% \\
                                             & Taiwan          & 193 & 77.7\% & 4.1\%  & 18.1\% \\
        \midrule
        Mergers                              & *               & 787 & 16.9\% & 8.3\%  & 74.8\% \\
        \midrule
        \textbf{Total}                       &                 & \textbf{1778}& \textbf{29.5\%} & \textbf{26.2\%} & \textbf{44.4\%} \\
        \bottomrule
    \end{tabularx}
    \caption{Number of examples (\#) and proportion of each stance label for each topic and stance target in the validation split
             of the dataset released by \citet{Liang_et_al_2024}. The stance labels are indicated by
             (\textbf{F})avour, (\textbf{A})gainst, (\textbf{N})eutral. The Mergers and Acquisitions topic covers 5 targets, which we combine here to save space.}
    \label{tab:dataset}
\end{table}

\subsubsection{Preprocessing}\hfill\\
In line with X's developer agreement\footnote{\url{https://developer.x.com/en/more/developer-terms/agreement-and-policy}}, the dataset released by \citet{Liang_et_al_2024} contains only post IDs. We therefore obtain the tweets and their paired images via the X developer API\footnote{\url{https://docs.x.com/x-api/introduction}}. All tweets are normalized and anonymized by replacing all URLs with the string \texttt{HTTPURL} and all user mentions with the string \texttt{@USER}. Following \citet{Liang_et_al_2024} and for fair comparison with their results, we use examples with videos and GIFs by extracting the first frame as an image.

\subsubsection{Translating the dataset}\hfill\\
The dataset as released contains English-language tweets only. In order to study the performance of VLMs across languages, we produce machine translations of this dataset into 6 additional languages: German (de), Spanish (es), French (fr), Hindi (hi), Portuguese (pt), and Chinese (zh). These languages were chosen to represent a range of language families (Germanic, Romance, other Indo-European, and Sino-Tibetan) and scripts (Latin, Devanagari, Simplified Chinese). A fair comparison across languages is paramount, so we opted to translate the dataset to obtain parallel texts --as opposed to sourcing disparate datasets in other languages-- in order to ensure model predictions are comparable across languages. We obtained translations using the Google Translate API.\footnote{\url{https://cloud.google.com/translate/docs/reference/rest/}} While it would be ideal to obtain human translations of the dataset to eliminate errors introduced by machine translation and remain faithful to the natural idiomatic variability that occurs across languages, this was not possible because human translation is a cumbersome and expensive task, that cannot be easily reproduced for multiple languages. Still, Google Translate has been shown in previous evaluations to perform well on these languages, and we maintain that the benefits of having parallel texts outweigh the potential pitfalls of using machine translation \cite{aiken2019updated,Taira_Kreger_Orue_Diamond_2021}. The result is 7 datasets: the original English dataset plus its translations into the 6 languages.

\subsection{Models}

We chose 4 state-of-the-art open-source VLMs to evaluate on the above datasets, detailed in \cref{tab:models}. We chose these particular models as they have similar overall model size and they provide a diverse sample of language model (LM) and vision model (VM) components. Additionally, the LM components of each model are either explicitly multilingual (e.g. Llama 3.1) or have demonstrated multilingual capabilities despite being advertised as English only (e.g., Gemma2). We provide a summary of each model's multilingual capabilities later in \cref{tab:language_support}.

\begin{table}[h]
    \centering
    \small
    \begin{tabular}{lrll}
        \toprule
        VLM           &  \# & LM             & VM             \\
        \midrule
        InternVL2     &  7B & InternLM2.5 7B & InternViT 300M \\
        Qwen2-VL      &  8B & Qwen2.5 7B     & ViT 675M       \\
        Ovis 1.6      &  9B & Gemma2 9B      & SigLIP 400M    \\
        Llama-Vision  &  11B& Llama 3.1 10B  & ViT 860M       \\
        \bottomrule
    \end{tabular}
    \caption{Vision Language Models (VLMs) evaluated ordered by total model size in number of parameters (\#) as well as their Language Model (LM) and Vision Model (VM) components.}
    \label{tab:models}
\end{table}

Our implementation uses Huggingface and Pytorch \cite{pytorch}.
%\footnote{The Huggingface paths for the models used are: \texttt{OpenGVLab/InternVL2-8B}, \texttt{Qwen/Qwen2-VL-7B-Instruct}, \texttt{AIDC-AI/Ovis1.6-Gemma2-9B}, \texttt{meta-llama/Llama-3.2-11B-Vision-Instruct}.}
In all experiments, we evaluate the models using 0-shot prediction on the validation split. We fix the random seed to 0 for all experiments to ensure reproducibility. All experiments were run on a single A100 GPU with 40GB of VRAM. The instruction prompt is always provided to the model in English no matter the language of the target dataset, as we found this to perform better overall than translating the prompt into the target language. The prompt template is given in \cref{fig:prompt}. 

\begin{figure}
    \centering
    \begin{subfigure}{\linewidth}
    \begin{verbbox}[\small\slshape]
  <|image|> From the image and tweet,    
  determine the stance regarding     
  {target}. The possible stance labels  
  are "favor", "against", or "neutral".   
  Answer with the label first, before   
  any explanation. Tweet: {tweet} \end{verbbox}
    \noindent\hspace{1em}\fbox{\theverbbox}
    \end{subfigure}\hfill\vspace{1em}
    \begin{subfigure}{\linewidth}
        \begin{verbbox}[\small\slshape]
  From the tweet and the text extracted
  from the image, determine
  the stance regarding {target}. The
  possible stance labels are "favor",
  "against", or "neutral". Answer with
  the label first, before any
  explanation. Tweet: {tweet},
  Image Text: {image_text} \end{verbbox}
    \noindent\hspace{1em}\fbox{\theverbbox}
    \end{subfigure}
    \caption{The instruction prompt templates for the Tweet \& Image (top) and Image Text (bottom) experiments. \texttt{<|image|>} is the special token whose embedding is set according to the vision model, \texttt{\{target\}} is the stance target for the given example, such as ``Joe Biden'' or ``Merger and acquisition between Aetna and Humana'', \texttt{\{tweet\}} is the tweet text, and \texttt{\{image\_text\}} is the plain text extracted from the image by the OCR tool.}
    \label{fig:prompt}
\end{figure}

\hfill

In the following sections we describe our three sets of experiments, which target multimodality, multilinguality, and their intersection, respectively.

\begin{figure*}[ht]
    \centering
    \begin{subfigure}{0.3\linewidth}
        \includegraphics[width=0.97\linewidth]{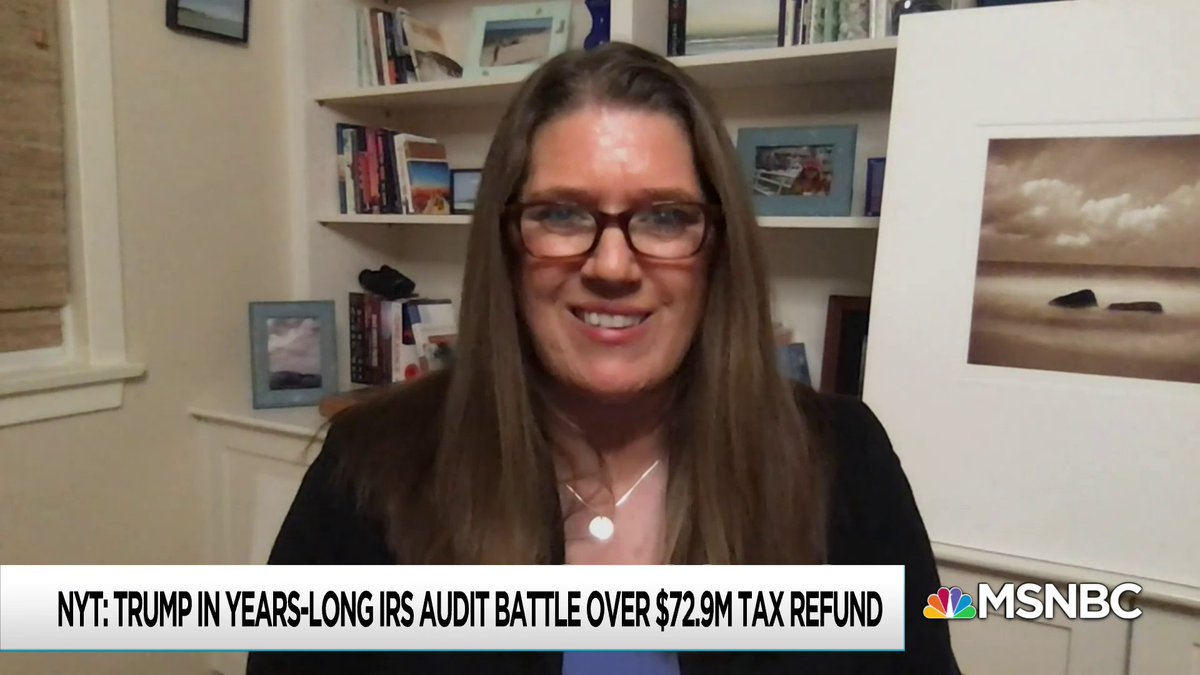}
    \end{subfigure}
    \hfill
    \begin{subfigure}{0.3\linewidth}
        \includegraphics[width=\linewidth]{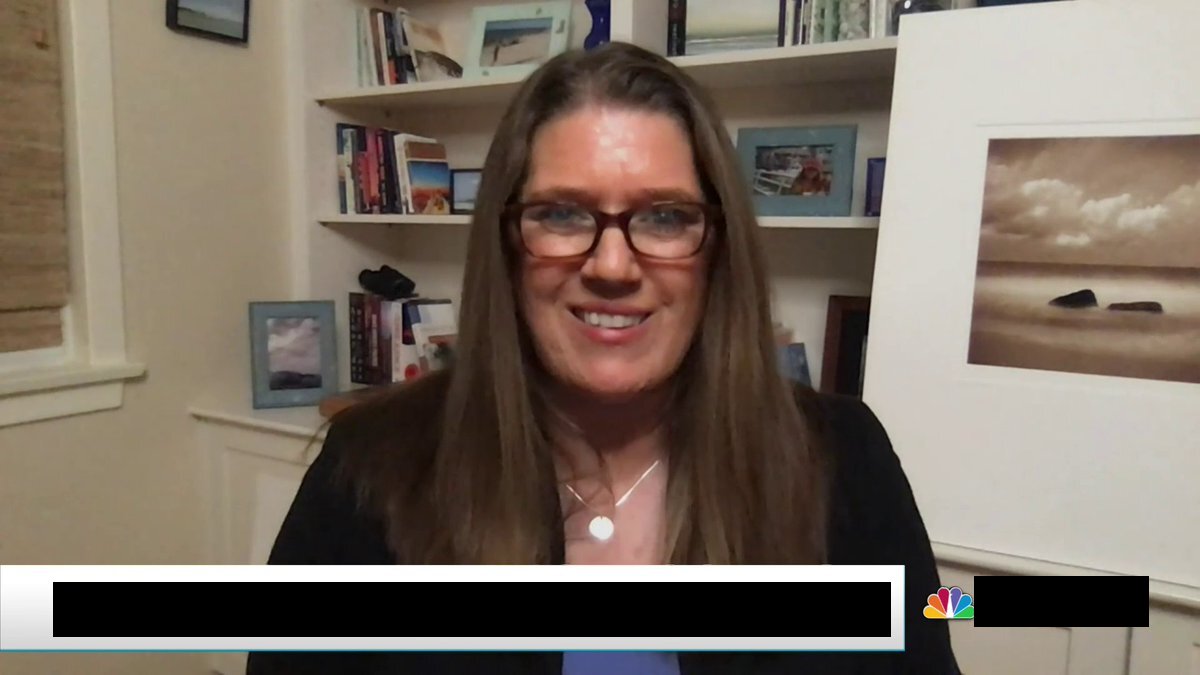}
    \end{subfigure}
    \hfill
    \begin{subfigure}{0.3\linewidth}
        \includegraphics[width=\linewidth]{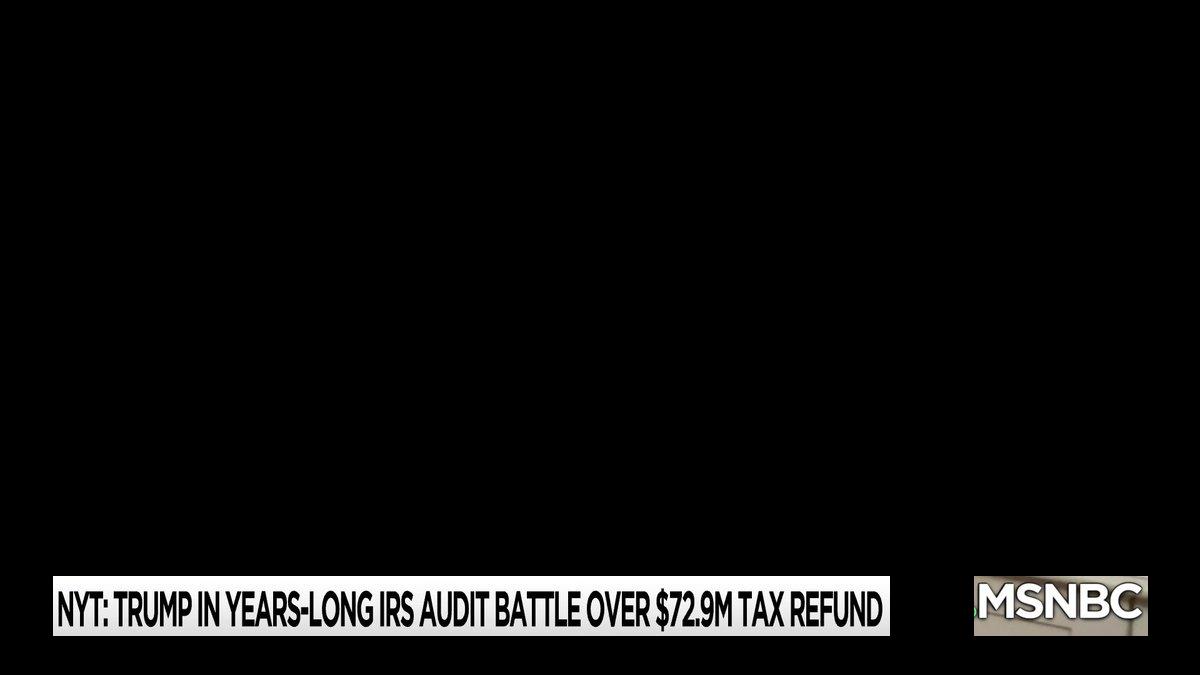}
    \end{subfigure}
    \begin{subfigure}{\linewidth}
        \centering
        \includegraphics[width=0.5\linewidth]{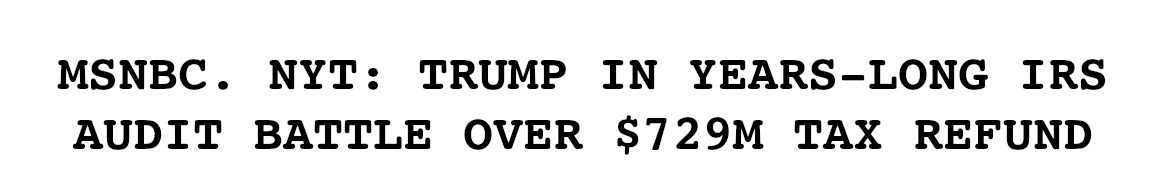}
    \end{subfigure}
    \caption{Left: An image which contains text that would be useful for prediction of the stance regarding Donald Trump. Middle: The same image after covering up the text using the bounding box output of the OCR tool, as used in the Text Blackout experiments. Right: Likewise for the Content Blackout experiments. Bottom: The plain text extracted by the OCR tool as used in the Image Text experiments.}
    \label{fig:blackout_example}
\end{figure*}

\subsection{Experiment 1: Multimodality}
\label{sec:exp1}

According to \citet{Liang_et_al_2024}, 46\% of images contained in the dataset convey stance information. It is thus crucial to determine whether the VLMs are actually leveraging information contained in the images to predict the stance. Therefore, the goal of this set of experiments is to determine the extent to which each model uses its vision component. We conduct two experiments. The first examines the overall contribution of the text and image modalities. The second delves deeper into which portions of the images are most useful for prediction. These evaluations are conducted on the original English dataset only, since this is free from any noise introduced by the machine translations and is the primary language for each model's LM component.

\subsubsection{Contribution of text and images}\hfill\\
First, it is important to establish the general contribution of the text and image modalities to the overall performance of the VLM. We therefore compute the performance in terms of macro F1 of each model in three scenarios:

\begin{itemize}
    \item \textbf{Tweet \& Image:} We input the instruction prompt containing the tweet text as well as the image. This is the default evaluation scenario.
    \item \textbf{Tweet only:} We input the instruction prompt containing the tweet text, but replace the image with Gaussian noise. This way the model is still forced to use the vision component but obtains no useful information from the image.
    \item \textbf{Image only:} We input the instruction prompt and the image, but leave the tweet text empty. I.e., the \texttt{\{tweet\}} variable in the instruction prompt is an empty string.
\end{itemize}

\noindent
We then compare the Tweet/Image Only macro F1 scores to that of the Tweet \& Image scenario. Statistical significance between predictions in each scenario is computed using McNemar's test, which determines whether the errors made by two systems on the same data points are equal, with a significant result indicating different error distributions \cite{McNemar_1947}. The results of this experiment will indicate the extent to which the text and image components on their own contribute to each model's overall performance.

\subsubsection{The role of text in images}\hfill\\
According to the analysis performed by the dataset authors, 24\% of the images contain text that is useful for making a correct stance prediction (cf. Table 2 of \citet{Liang_et_al_2024}). Applying the GATE OCR tool\footnote{\url{https://cloud.gate.ac.uk/shopfront/displayItem/ml-ocr}} to the images, we further found that 67\% contain text of any sort. The VLMs we evaluate have demonstrated OCR capabilities (cf. \citet{Wang_et_al_2024} appendix A), suggesting that they should be able to obtain useful information from text contained in the images. We therefore investigate whether the models' VM components are able to leverage this in-image text when making a prediction. To test this we introduce three additional experiment scenarios, which ablate certain portions of the images.

\begin{itemize}
    \item \textbf{Text Blackout:} Using the bounding boxes detected by the GATE OCR tool, we cover all text in each image with a black box. In other words, we remove the text cues from the images while retaining all other visual information. 
    \item \textbf{Content Blackout:} This scenario is the inverse of Text Blackout. Instead of covering up all text in the image, we cover up all \textit{but} the text. The goal of this experiment is to isolate the text from all other visual information.
    \item \textbf{Image Text:} We extract the plain text from the images using the OCR and craft a new prompt that provides the models with both the tweet and the image text, shown in \cref{fig:prompt}. Like Content Blackout, this isolates the text cues from other visual cues but also bypasses the visual encoder entirely. In this case, the VM is only ever given Gaussian noise.
\end{itemize}

In each scenario, we evaluate the models in the Image Only and Tweet \& Image scenarios, using the blacked out image or the plain text extracted from the image. Comparing the results from each to the original, unmodified Image Only and Tweet \& Image results will provide insight into how the visual component of each model is using the content of the images. Additionally, comparing the Image Text experiments to the Content Blackout experiments specifically will tell how how effectively the visual components are leveraging text contained in the image. Examples of the model inputs in each scenario are given in \cref{fig:blackout_example}.

Finally, we further explore the effect of in-image text on model predictions by estimating a logistic regression model to predict whether a VLM's prediction changes from correct to incorrect after blacking out portions of the image. The resulting regression coefficients will show whether, on average across the dataset, removing a portion of the image is correlated with a model changing a correct prediction to an incorrect one, which would suggest that content is helpful for making correct predictions.

\subsection{Experiment 2: Multilinguality}
\label{sec:exp2}

As discussed in the introduction, information online often spreads across geographic regions and languages. Given that the LM components of many VLMs demonstrate multilingual capabilities, it is important to determine whether prediction performance is consistent across languages given input that is semantically identical. Thus, in this set of experiments we compare the performance of each model across languages, using the translated datasets to control for semantic content. We first conduct a brief evaluation regarding the extent to which each VLM supports each language. Then, we perform a comparative model evaluation that investigates the following two ways in which predictions may differ.

\subsubsection{Performance across languages}\hfill\\
As in Experiment 1, we compare the macro F1 score of each model across the dataset languages in the Tweet Only, Image Only, and Tweet \& Image scenarios, and perform a McNemar's test between predictions in each language and the English results to determine whether there is a statistically significant difference in the prediction errors.\footnote{Because the prompt is always in English and the translated tweet is not provided, we expect the performance in the Image Only scenario to be identical across languages.}

\subsubsection{Agreement between languages}\hfill\\
A model may have similar performance in two languages as measured by F1 score but high disagreement at an instance level. For example, two evaluations may have similar numbers of a certain class predicted (in)correctly, but the precise examples that they predict (in)correctly don't overlap. In essence, we are interested in measuring the extent to which a model agrees with itself when the tweet language changes. We measure this by computing the Cohen's kappa between the predictions in each language for each model. Because the inputs in two languages are semantically identical and their label distributions are the same, we would expect that a model ``fluent'' in each language would produce exactly the same labels for each language and produce random labels for any language it does not understand.
 
\subsection{Experiment 3: The Intersection of Multimodality and Multilinguality}
\label{sec:exp3}

Experiments 1 and 2 investigate modality and language separately. Here, we investigate how the impact of input language on each VLM's reliance on the text and image modalities. To do this we conduct the following experiments.

\begin{itemize}
    \item {Because the image input is language-independent, the amount of information it is able to provide to the model is the same for each language (this is confirmed later in \cref{fig:per_language_performance}). We therefore investigate the contribution of the vision modality on top of the tweet by measuring the difference in F1 between the Tweet \& Image and Tweet Only scenarios for each language. A larger positive difference suggests a greater reliance on the image input.}
    \item {To investigate the effect of in-image text vs. other image content, we estimate a logistic regression model for inputs in each language besides English as we did in Experiment 1. We then compare the resulting regression coefficients to those for English. A difference in coefficients suggests the model is more or less reliant on that type of image content for the target language.}
\end{itemize}

\section{Results}

We here present and discuss the results of each set of experiments in turn. We provide a high-level summary of our findings as well as key takeaways later in the Discussion section.

\subsection{Experiment 1: Multimodality}

\begin{figure}[t]
    \centering
    \includegraphics[width=\linewidth]{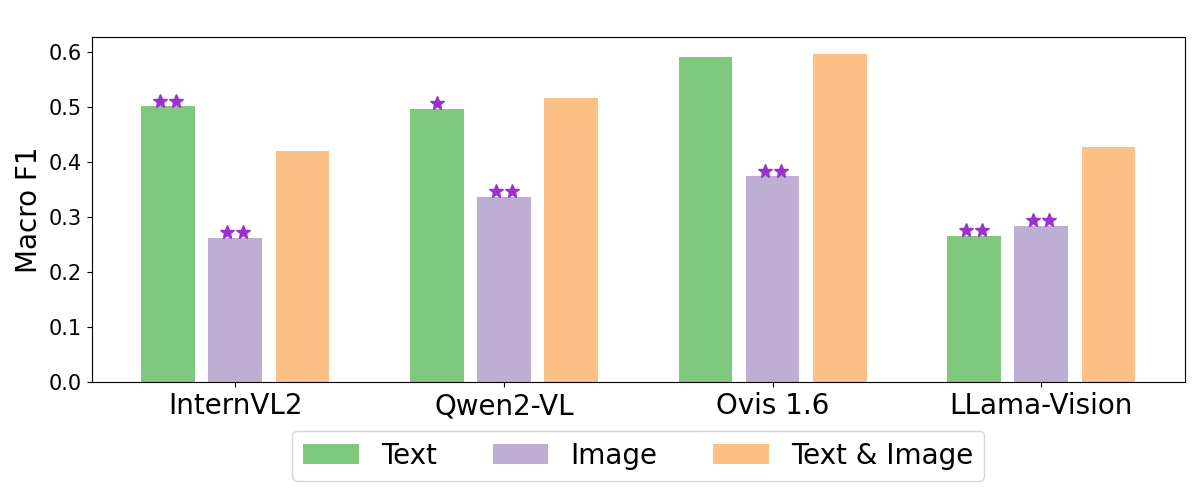}
    \caption{Performance of each VLM on the English dataset for each evaluation scenario. Statistical significance vs. Tweet \& Image indicated as * p $\leq$ 0.05, ** p $\leq 0.005$.}
    \label{fig:en_performance}
\end{figure}

\cref{fig:en_performance} shows the performance of each VLM in the Tweet Only, Image Only, and Tweet \& Image scenarios on the English data. Three of the four models (InternVL2, Qwen2, and Ovis) perform significantly better in the Tweet Only and Tweet \& Image scenarios than the Image Only scenario. This is intuitive, as the analysis in \citet{Liang_et_al_2024} found that only 46\% of images contain information relevant to the stance label. The exception is Llama-Vision, which performs the worst in the Tweet Only scenario and the worst overall among the models, despite having the greatest number of parameters at 11B. Llama-Vision does, however, have the largest vision model of those evaluated, so it may be that it is more reliant on the image modality. It is also interesting to note that for InternVL2, inclusion of the image modality \textit{hurts} performance over the Tweet Only scenario, where for all other models it either improves or does not significantly change the results. InternVL2 does have the smallest vision model at 300M parameters and performs worst overall in the Image Only scenario, so it may be that its vision model is simply not powerful enough to encode useful information from the images.

A more in-depth investigation of each model's use of each modality is given in \cref{tab:exp1_blackout_img_text}, which shows results for the image ablation experiments. To better understand the role of text and other content in the images, we split the evaluation according to whether the image does or does not contain text according to the output of the OCR tool.

\begin{table*}[]
    \centering
    \begin{tabular}{l|llllllll}
        \toprule
                            & \mc{2}{c}{InternVL2} & \mc{2}{c}{Qwen2-VL} & \mc{2}{c}{Ovis 1.6} & \mc{2}{c}{Llama-Vision}   \\
                            &\tc{T}   & \tc{N}  &\tc{T}   & \tc{N}  & \tc{T}   & \tc{N}&\tc{T}   & \tc{N} \\
        \midrule
        \textbf{Tweet Only}  & 0.518   & 0.452   & 0.521   & 0.432   & 0.597   & 0.565  & 0.266   & 0.264  \\
        \midrule
         \textbf{Image Only}& 0.277   & 0.190   & 0.370   & 0.246   & 0.409   & 0.286  & 0.294   & 0.261  \\
         v. Text Blackout   & 0.176** & 0.190   & 0.187** & 0.246   & 0.313** & 0.286  & 0.260   & 0.262  \\
         v. Content Blackout& 0.257** & 0.129** & 0.318** & 0.127** & 0.352** & 0.099**& 0.274   & 0.149**\\
         v. Image Text      & 0.252** & 0.120** & 0.218** & 0.114** & 0.285** & 0.093**& 0.202** & 0.130**\\
         \bottomrule\toprule
      \textbf{Tweet \& Image}& 0.394   & 0.462   & 0.527   & 0.479   & 0.614   & 0.563  & 0.441   & 0.379  \\
         v. Text Blackout   & 0.444** & 0.462   & 0.527   & 0.477   & 0.621   & 0.563  & 0.376** & 0.378  \\
         v. Content Blackout& 0.400   & 0.493   & 0.508   & 0.423*  & 0.611   & 0.570  & 0.419   & 0.364* \\
         v. Image Text      & 0.392   & 0.461   & 0.452** & 0.433** & 0.585*  & 0.560  & 0.332** & 0.290**\\
        \bottomrule
    \end{tabular}
    \caption{Comparison of macro F1 scores on the English data for the Blackout and Image Text scenarios vs their unmodified counterparts. Subcolumns under each model indicate results on examples with text (T) and without text (N) in the images. Statistical significance vs. the Image Only or Tweet \& Image row is indicated as * p $\leq$ 0.05, ** p $\leq 0.005$.}
    \label{tab:exp1_blackout_img_text}
\end{table*}

\begin{table}[ht!]
    \small
    \centering
    \begin{tabular}{cl|llll}
        \toprule
               & Blackout& Intern    &  Qwen      &  Ovis     & Llama     \\
        \midrule
\mr{2}{*}{I}   & Text    & 0.546**   &  0.549**   &  0.560**  &  0.546   \\
               & Content & 0.509**   &  0.508**   &  0.514**  &  0.535   \\
       \midrule
\mr{2}{*}{T\&I}& Text    & 0.523**   &  0.525**   &  0.526*   &  0.558*  \\
               & Content & 0.512**   &  0.510*    &  0.510    &  0.548   \\
       \bottomrule
    \end{tabular}
    \caption{Probability of each model changing a correct prediction to an incorrect one after removing text or other content from the images. I indicates results for the Image Only scenario. T\&I indicates results for the Tweet \& Image scenario. Statistical significance of the regression coefficient used to compute the probability indicated as * p $\leq$ 0.05, ** p $\leq 0.005$.}
    \label{tab:exp1_lr}
\end{table}

We first discuss results in the Image Only scenario. We note that removing any visual information (i.e., Text or Content Blackout) reduces performance. However, on those images that contain text (T columns), blacking out the text portion of the image has a greater negative effect on the F1 score than blacking out other content. For example, Qwen2's F1 drops by 0.18 when blacking out the text, but only 0.05 when blacking out other image content. This suggests that when text is present in the image, it is helpful for making a correct prediction. We also notice that using the plain image text only performs worse than Content Blackout for all models, suggesting that the visual organization of the text in the image is important, since this is the only aspect that is lost between the Image Text and Content Blackout experiments.

Trends are less clear in the Tweet \& Image scenario. Only a few results are significant and those that are occasionally go against intuition. For example, InternVL2 performs \textit{better} after blacking out the image text and for Qwen2 and Ovis blacking out Text or Content does not significantly change model performance. This may suggest that these models are more reliant on the text modality when it is available. Nevertheless, these models do obtain some information from the image modality, as indicated by a significant performance decrease when using only the plain text extracted from the images (i.e., the Image Text rows). Like in the Image Only scenario, Image Text underperforms Content Blackout across models on examples that contain in-image text, suggesting that the visual organization of the text is important.

We can gain more insight into the role of text in the images by controlling for the proportion of the images that contain text, as indicated by the areas of the bounding boxes identified by the OCR tool. To determine the effect of text coverage on performance, we estimate a logistic regression model with text coverage as the independent variable to predict whether the VLM's predictions changed from correct to incorrect after blacking out the text or other content in the images. The resulting coefficients for the intercept and independent variable are then used to compute the probability that each model would change a correct prediction to an incorrect one after blacking out a portion of the image.\footnote{We compute these probabilities for a hypothetical image with equal proportions of text and other content.} These probabilities are shown in \cref{tab:exp1_lr}.

These probabilities show similar trends to those given in \cref{tab:exp1_blackout_img_text}. In the Image Only scenario, there is a higher probability for each model to change a correct prediction to an incorrect one after blacking out the text over blacking out content. This mirrors the larger drop in F1s reported in \cref{tab:exp1_blackout_img_text} after blacking out text. Again, Llama-Vision is an exception to this trend: the coefficients estimated by the logistic regression were insignificant as were the differences in F1s in the Text and Content Blackout experiments in \cref{tab:exp1_blackout_img_text}. Also, where changes in F1s were somewhat inconsistently significant in the Tweet \& Image scenario in \cref{tab:exp1_blackout_img_text}, the regression coefficients are significant in all but three cases in \cref{tab:exp1_lr}. This suggests that in both the Image Only and Tweet \& Image scenarios the in-image text content is generally more important for prediction than other content.  

Overall, these results alongside those from \cref{tab:exp1_blackout_img_text} indicate that InternVL2, Qwen2-VL, and Ovis 1.6 are indeed using in-image text to a greater extent than other content when making predictions. This is, \textit{despite the fact that the images on average have only 28\% of their total area occupied by text}, according to the bounding boxes identified by the OCR tool.

\begin{figure*}[h]
    \centering
    \includegraphics[width=\linewidth]{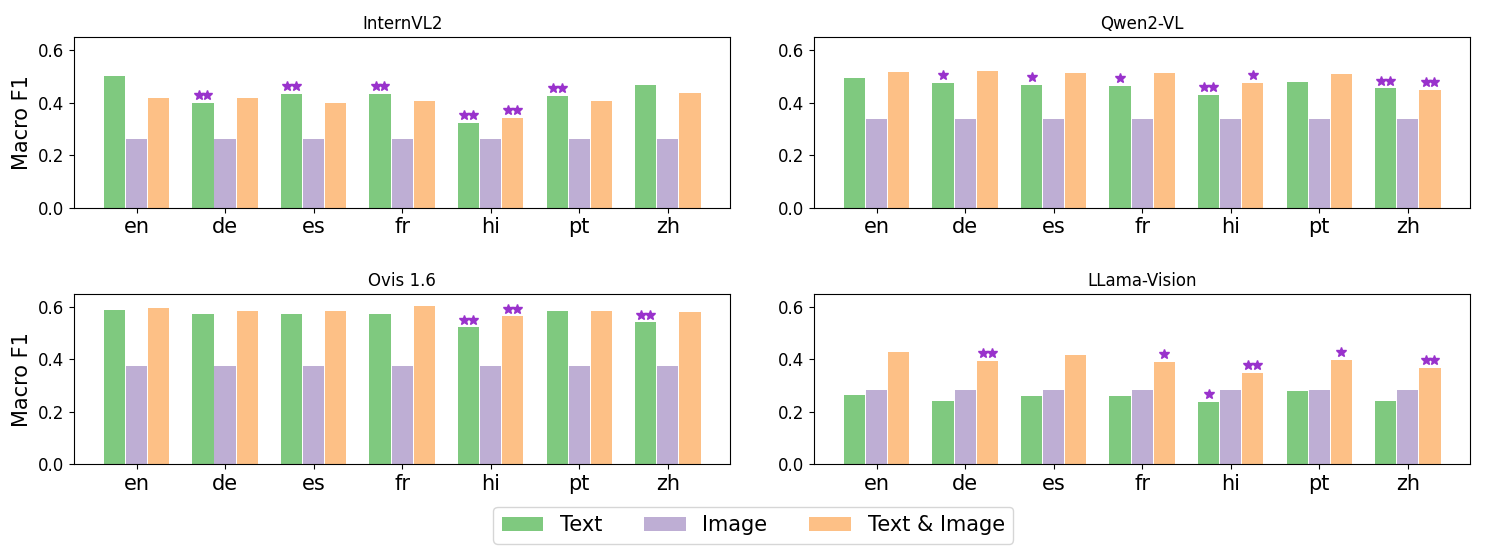}
    \caption{Macro F1 of each VLM on each language in each evaluation scenario. Statistical significance vs. the corresponding English results is computed using a McNemar's test and is indicated as * p $\leq$ 0.05, ** p $\leq 0.005$.}
    \label{fig:per_language_performance}
\end{figure*}

\begin{table}[ht!]
    \centering
    \begin{tabular}{lccccccc}
        \toprule
        VLM           & en & de & es & fr & hi & pt & zh \\
        \midrule
        InternVL2     &\ck & 2/5& 3/5& 2/5& 0/5& 3/5& 5/5\\
        Qwen2-VL      &\ck &\ck &\ck &\ck & 3/5&\ck &\ck \\
        Ovis 1.6      &\ck & 4/5& 4/5& 4/5& 5/5& 5/5& 5/5\\
        Llama-Vision  &\ck &\ck &\ck &\ck &\ck &\ck & 3/5\\
        \bottomrule
    \end{tabular}
    \caption{Languages supported by each VLM. A \ck indicates that the model officially supports that language according to its documentation. For languages that are not officially supported, fractions indicate the number of tasks for which the given model demonstrated understanding and generation capabilities in our evaluation.}
    \label{tab:language_support}
\end{table}

\subsection{Experiment 2: Multilinguality}

We first attempt to establish the extent to which each VLM supports each language beyond what each is reported to officially support in the documentation. Towards this, we performed an evaluation of each model's performance on each language for 5 tasks: question answering, target language to English translation, English to target language translation, image description, and image question answering. We manually evaluated the output of each model on each task according to whether the output was fluent, and in the target language. The results are given in \cref{tab:language_support}, and additional details on these experiments are given in the appendix.
We see that all models support all of our target languages to some extent, with the only exception being InternVL2 which did not demonstrate any understanding of Hindi.

The F1 scores of each model across languages in each evaluation scenario are given in \cref{fig:per_language_performance}. Bars marked with stars indicate results that are significantly different from the corresponding English results according to a McNemar's test.
We can determine the consistency of each model across languages by examining the F1 discrepancy compared to English as well as the number of languages whose predictions differ significantly from English. Overall, we see that Ovis is the most consistent across languages as its F1 scores are all similar and only two languages (Hindi and Chinese) have predictions that differ significantly from English. While Qwen2 has similar F1 scores across languages, the McNemar's tests revealed that its prediction errors for five out of six languages differ significantly from English. InternVL2 and Llama-Vision also differ significantly for five languages and their F1 scores are less consistent. Again we see that Llama-Vision performs poorly despite being the largest model evaluated: in addition to being the worst performing model overall it is also one of the least consistent across languages.

\begin{figure*}[h]
    \centering
    \includegraphics[width=\linewidth]{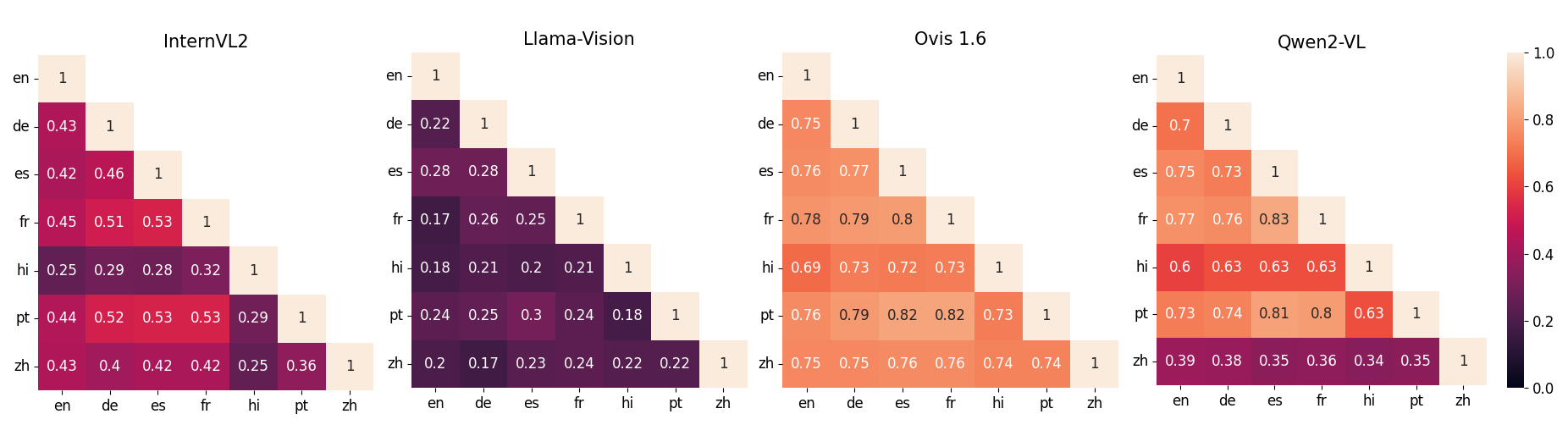}
    \caption{Cohen's kappa values between predictions for each pair of languages  in the Text \& Image scenario. The Text/Image Only scenarios exhibit the same trends so are not pictured for space reasons.}
    \label{fig:text_image_kappas}
\end{figure*}

We further investigate the consistency of each model's predictions across languages in \cref{fig:text_image_kappas}, which shows the Cohen's kappa between predictions in each pair of languages. First, we note that agreement seems to correlate with performance, with better performing models achieving higher kappa scores. Again we see that Ovis is the most consistent, with kappa scores $\geq 0.7$ between all languages, and that Llama-Vision is the least consistent with all scores $\leq 0.3$. Hindi is the least consistent across all models, although this is somewhat expected as it is only officially supported by Llama-Vision (cf. \cref{tab:language_support}). Additionally, there is notable disagreement between Chinese and all other languages on Qwen2. This is particularly interesting because Qwen2 was developed with a specific focus on Chinese and English, and officially supports 6 of the 7 languages investigated here \cite{Wang_et_al_2024}.

\begin{figure}[ht!]
    \centering
    \includegraphics[width=\linewidth]{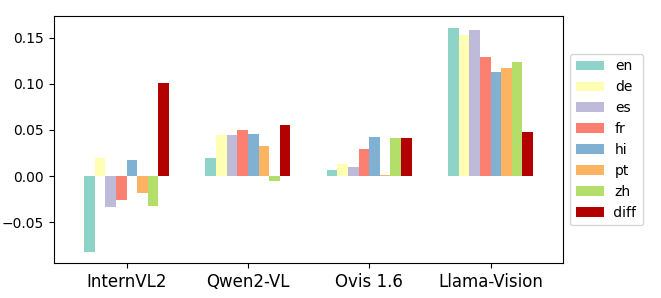}
    \caption{Differences in macro F1 between the Tweet \& Image and Tweet Only scenarios for each model and language. Positive values indicate Tweet \& Image performance greater than Tweet Only and \textit{vice versa} for negative values. The dark red bar labeled ``diff'' indicates the variability of that model across languages, computed as the absolute difference between the maximum and minimum values for each language.}
    \label{fig:lang_trends}
\end{figure}

\subsection{Experiment 3: Joint Multimodality and Multilinguality}

As a first step towards understanding interactions between modality an language across models, we reexamine the results for each language between Text Only, Image Only, and Text \& Image scenarios given in \cref{fig:per_language_performance}. Because the Image Only scenario is independent of the dataset language (i.e., the results are identical across languages), we can view the difference between the Tweet \& Image and Tweet Only F1 on each language as a proxy for the amount of information contributed by the vision component.

The results of this comparison are shown in \cref{fig:lang_trends}. Examining the bars for each language, we see that all models besides InternVL2 show a positive reliance on the vision modality (with Qwen2 on Chinese being a notable exception). For InternVL2 we see that the vision modality generally \textit{harms} performance, although the harm is greatest for English. For Qwen2 and Ovis there is in general a greater positive effect of the vision modality for languages other than English, while the opposite is true for Llama-Vision.

The dark red bar in \cref{fig:lang_trends} is a measure of variability, computed as the absolute difference between the maximum and minimum values of the other bars for a given model. It can thus be viewed as a measure of consistency in the effect of the vision modality across languages. From these red bars, we see that Ovis is the most consistent overall in its use of the vision modality, followed by Llama-Vision and Qwen2. While the variability is greatest for InternVL2, we note that because it expresses a fairly consistent \textit{decrease} in F1 from Tweet Only to Tweet \& Image, it is difficult to conclude the extent to which it relies on the vision modality for each language. 

\begin{figure*}
    \centering
    \includegraphics[width=0.85\linewidth]{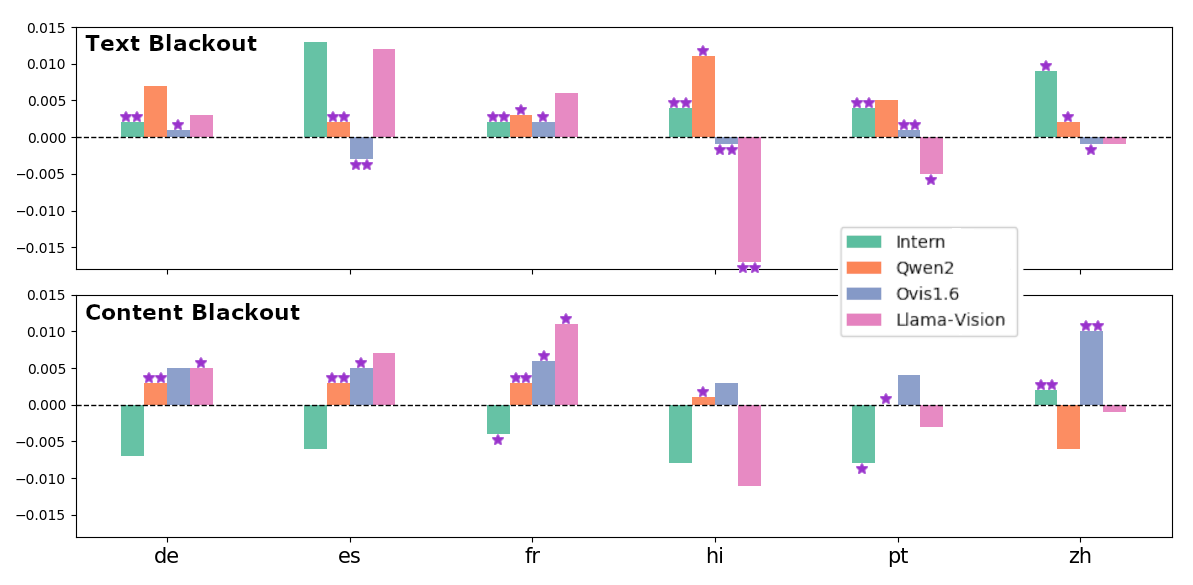}
    \caption{Differences in probability of a prediction becoming incorrect after Text or Content Blackout for each model-language pair vs. English in the Text \& Image scenario, computed using the logistic regression coefficients. Statistical significance of the corresponding regression coefficient is indicated as * p $\leq$ 0.05, ** p $\leq 0.005$.}
    \label{fig:lr_prob_diffs}
\end{figure*}

To investigate the role of text in the images for languages other than English, we proceed as in Experiment 1, estimating logistic regression models for the other languages and using the coefficients to compute the probability of each model changing a correct prediction to an incorrect one after removing in-image text or other content. We report the differences in probabilities vs. the English results (cf. \cref{tab:exp1_lr}) for each model-language pair in \cref{fig:lr_prob_diffs}. These results only cover the Tweet \& Image scenario since we are specifically interested in how the tweet language affects the models' use of the image modality.

\cref{fig:lr_prob_diffs} can be interpreted as follows. A significant \textit{positive} difference indicates that the model is \textit{more} likely to flip a correct prediction for the given language than it is for English, while a significant \textit{negative} difference indicates this is \textit{less} likely. A positive difference further suggests that the model is more reliant on what was blacked out on the target language vs. English, and \textit{vice versa} for a negative difference.

Overall, the differences are small, in the range $\pm0.015$ for all models and languages, which suggests that the models are generally consistent regarding their reliance on the in-image text and other content across languages. It is notable that InternVL2 and Llama-Vision tend to have more extreme differences than Ovis and Qwen2, for example on Spanish and Hindi after blacking out the text and on Hindi after blacking out content. Of particular interest is Llama-Vision on Hindi in the Text Blackout scenario. According to the figure we see that Llama-Vision is much less likely to change a correct to an incorrect prediction in Hindi than in English, as there is a significant difference in probability of -0.017. The F1 scores support this: on English Llama-Vision drops 0.065 F1 (0.441 $\rightarrow$ 0.376) after blacking out text but only drops 0.038 on Hindi (0.349 $\rightarrow$ 0.311), suggesting that Llama-Vision relies more on the in-image text when the tweet is in Hindi than when it is in English.

\cref{fig:lr_prob_diffs} also provides additional context to the consistency results from Experiment 2. Ovis, as the most consistent model across languages in terms of F1 score, is also highly consistent with English across languages after blacking out the text, while it generally relies more on the non-text content of the images in languages other than English. On the other hand, Qwen2 is much more consistent when blacking out content and more reliant on the text in the images on languages besides English.

\section{Discussion}

In summary, we identified the following takeaways:

\begin{enumerate}
    \item {The VLMs are more reliant on the text modality than the image modality for this task, with inclusion of the image modality providing only small performance increases over text alone (cf. \cref{fig:en_performance}). The exception is Llama-Vision, which is generally more reliant on the vision modality than text.}
    \item {When provided with images, VLMs rely more on the text contained in the images than the other content, as suggested by the performance drops in \cref{tab:exp1_blackout_img_text} and differences in probabilities in \cref{tab:exp1_lr}. This is despite text occupying a minority portion of most images. Additionally, our experiments using plain text extracted from the images suggest that the visual organization of the text in the images is important, rather than only the text itself.}
    \item {The VLMs produce generally consistent predictions across languages, although with some idiosyncrasies. These include performance drops on Hindi across models (cf. \cref{fig:per_language_performance}) and Qwen2's inconsistent predictions on Chinese (cf. \cref{fig:text_image_kappas}). Despite officially supporting English only, we found Ovis to be the most consistent across the languages tested with its predictions differing significantly from English for only 2 languages. It is also the best performing model overall, obtaining a macro F1 score of 0.614 in the Text \& Image scenario.}
    \item {Each model exhibits a relatively consistent reliance on the text and vision modalities across languages, evidenced by similar per-language trends in F1s across the Text Only, Image Only, and Text \& Image scenarios in \cref{fig:per_language_performance}. When it comes to reliance on the in-image text and other content, Ovis is again highly consistent, yet we see a slight increased reliance on non-text content on languages other than English.}
\end{enumerate}

The results emphasize that the largest model is not necessarily the best performing (Llama-Vision), and just because a VLM claims to support multiple languages, does not necessarily mean that its predictions will be consistent across those languages (Llama-Vision and Qwen2-VL, especially on Chinese). Further, the high consistency exhibited by Ovis 1.6 suggests that its LLM, Gemma2, is in practice multilingual, despite claiming to support English only.

\subsection{Limitations and Future Work}

Regarding our overall evaluation methodology, we used a fixed random seed for all experiments in order to ensure reproducibility and because we did not have the computational resources to run each experiment over multiple seeds. However, we noticed anecdotally that the labels predicted by the models were somewhat variable depending on the random seed, so in the future it would be worthwhile to rerun all experiments with different seeds to determine whether this has any effect on the results. In addition, we only used one dataset since it was the only stance detection dataset with paired images and text available at the time our research began. Future work should extend this analysis to new multimodal stance datasets such as MultiClimate \cite{Wang_Zuo_Peng_Plank_2024} (released after our research had begun) and may require the creation of novel datasets in this area.

\paragraph{Modality Limitations}\hfill\\
A number of examples in the dataset used here contain videos or GIFs. We followed the dataset authors by using the first frame of these examples for our evaluation, but this certainly resulted in a loss of information. Future work ought to leverage some VLMs' (e.g., Qwen2's) ability to process different types of vision inputs in order to realize the dataset's full potential.

This study investigated VLMs' reliance on modalities by analysing their overall prediction distributions. Future explorations might obtain more instance-level insights by using attention distributions or feature attribution methods to 1) quantify the effect of the text and vision modalities on single examples and 2) investigate interactions (e.g., in terms of attention weights) between text and images.

\paragraph{Language Limitations}\hfill\\
As previously mentioned, using Google Translate to obtain translations of the dataset is not ideal and may have introduced errors into the dataset that we could not account for in our evaluations. Additionally, we avoided translating the dataset into low-resource languages as the quality of Google Translate on these can be poor \cite{aiken2019updated}. Future work on evaluating the multilingual capabilities of these models ought to obtain human translations and ideally introduce lower-resource languages as well as additional language families/scripts. VLMs are expected to perform worse on these languages which may highlight inconsistencies in predictions that were not noticeable here. Finally, it was too resource-intensive for us to perform a full evaluation of each VLM's language support --covering a large variety of tasks and examples-- so our evaluation in \cref{tab:language_support} was cursory.

\section{Conclusion}

This study explored the performance of four VLMs --InternVL2, Qwen2-VL, Ovis 1.6, and Llama-Vision-- on a stance detection task in terms of their reliance on the text and image modalities as well as performance variability on languages other than English. We found that VLMs are highly reliant on text for this task, both from the text modality and text contained in the image modality. This latter finding is surprising because of those images that contain any text at all, the text occupies on average only 28\% of their total area. Additionally, the models' predictions and reliance on the image modality are relatively consistent across the languages studied. Ovis, which officially only supports English, was notably the most consistent among the VLMs evaluated, and Llama-Vision, which officially supports all but one language studied, was found to be one of the least consistent.

\section{Acknowledgments}
This work is funded by the European Media and Information Fund (ExU - grant number 291191).
The sole responsibility for any content supported by the European Media and Information Fund lies
with the author(s) and it may not necessarily reflect the positions of the EMIF and the Fund Partners,
the Calouste Gulbenkian Foundation and the European University Institute.

\bibliography{refs}

\begin{thebibliography}{36}
\providecommand{\natexlab}[1]{#1}

\bibitem[{Aiken et~al.(2019)}]{aiken2019updated}
Aiken, M.; et~al. 2019.
\newblock An updated evaluation of Google translate accuracy.
\newblock \emph{Studies in linguistics and literature}, 3(3): 253--260.

\bibitem[{ALDayel and Magdy(2021)}]{ALDayel_Magdy_2021}
ALDayel, A.; and Magdy, W. 2021.
\newblock Stance detection on social media: State of the art and trends.
\newblock \emph{Information Processing \& Management}, 58(4): 102597.

\bibitem[{Alhindi et~al.(2021)Alhindi, Alabdulkarim, Alshehri, Abdul-Mageed, and Nakov}]{Alhindi_et_al_2021}
Alhindi, T.; Alabdulkarim, A.; Alshehri, A.; Abdul-Mageed, M.; and Nakov, P. 2021.
\newblock AraStance: A Multi-Country and Multi-Domain Dataset of Arabic Stance Detection for Fact Checking.
\newblock In Feldman, A.; Da~San~Martino, G.; Leberknight, C.; and Nakov, P., eds., \emph{Proceedings of the Fourth Workshop on NLP for Internet Freedom: Censorship, Disinformation, and Propaganda}, 57–65. Online: Association for Computational Linguistics.

\bibitem[{Barriere, Jacquet, and Hemamou(2022)}]{barriere-etal-2022-cofe}
Barriere, V.; Jacquet, G.~G.; and Hemamou, L. 2022.
\newblock {C}o{FE}: A New Dataset of Intra-Multilingual Multi-target Stance Classification from an Online {E}uropean Participatory Democracy Platform.
\newblock In He, Y.; Ji, H.; Li, S.; Liu, Y.; and Chang, C.-H., eds., \emph{Proceedings of the 2nd Conference of the Asia-Pacific Chapter of the Association for Computational Linguistics and the 12th International Joint Conference on Natural Language Processing (Volume 2: Short Papers)}, 418--422. Online only: Association for Computational Linguistics.

\bibitem[{Ceylan, Anderson, and Wood(2023)}]{Ceylan_Anderson_Wood_2023}
Ceylan, G.; Anderson, I.~A.; and Wood, W. 2023.
\newblock Sharing of misinformation is habitual, not just lazy or biased.
\newblock \emph{Proceedings of the National Academy of Sciences}, 120(4): e2216614120.

\bibitem[{Cruickshank and Ng(2023)}]{cruickshank2023use}
Cruickshank, I.~J.; and Ng, L. H.~X. 2023.
\newblock Use of large language models for stance classification.
\newblock \emph{arXiv preprint arXiv:2309.13734}.

\bibitem[{Devlin et~al.(2019)Devlin, Chang, Lee, and Toutanova}]{Devlin_Chang_Lee_Toutanova_2019}
Devlin, J.; Chang, M.-W.; Lee, K.; and Toutanova, K. 2019.
\newblock BERT: Pre-training of Deep Bidirectional Transformers for Language Understanding.
\newblock In \emph{Proceedings of the 2019 Conference of the North American Chapter of the Association for Computational Linguistics: Human Language Technologies, Volume 1 (Long and Short Papers)}, 4171–4186. Minneapolis, Minnesota: Association for Computational Linguistics.

\bibitem[{Dosovitskiy et~al.(2021)Dosovitskiy, Beyer, Kolesnikov, Weissenborn, Zhai, Unterthiner, Dehghani, Minderer, Heigold, Gelly, Uszkoreit, and Houlsby}]{Dosovitskiy_et_al_2021}
Dosovitskiy, A.; Beyer, L.; Kolesnikov, A.; Weissenborn, D.; Zhai, X.; Unterthiner, T.; Dehghani, M.; Minderer, M.; Heigold, G.; Gelly, S.; Uszkoreit, J.; and Houlsby, N. 2021.
\newblock An Image is Worth 16x16 Words: Transformers for Image Recognition at Scale.
\newblock (arXiv:2010.11929).
\newblock ArXiv:2010.11929 [cs].

\bibitem[{Elfardy and Diab(2016)}]{elfardy2016cu}
Elfardy, H.; and Diab, M. 2016.
\newblock Cu-gwu perspective at semeval-2016 task 6: Ideological stance detection in informal text.
\newblock In \emph{Proceedings of the 10th international workshop on semantic evaluation (SemEval-2016)}, 434--439.

\bibitem[{Fajcik, Smrz, and Burget(2019)}]{fajcik-etal-2019-fit}
Fajcik, M.; Smrz, P.; and Burget, L. 2019.
\newblock {BUT}-{FIT} at {S}em{E}val-2019 Task 7: Determining the Rumour Stance with Pre-Trained Deep Bidirectional Transformers.
\newblock In May, J.; Shutova, E.; Herbelot, A.; Zhu, X.; Apidianaki, M.; and Mohammad, S.~M., eds., \emph{Proceedings of the 13th International Workshop on Semantic Evaluation}, 1097--1104. Minneapolis, Minnesota, USA: Association for Computational Linguistics.

\bibitem[{Gürer, Hubbard, and Bohon(2023)}]{Gürer_Hubbard_Bohon_2023}
Gürer, D.; Hubbard, J.; and Bohon, W. 2023.
\newblock Science on social media.
\newblock \emph{Communications Earth \& Environment}, 4(1): 1–5.

\bibitem[{Hu et~al.(2023)Hu, Liu, Wang, Zhang, and Lin}]{Hu_Liu_Wang_Zhang_Lin_2023}
Hu, M.; Liu, P.; Wang, W.; Zhang, H.; and Lin, C. 2023.
\newblock MSDD: A Multimodal Language Dateset for Stance Detection.
\newblock In Su, Q.; Xu, G.; and Yang, X., eds., \emph{Chinese Lexical Semantics}, 112–124. Cham: Springer Nature Switzerland.
\newblock ISBN 978-3-031-28953-8.

\bibitem[{Kawintiranon and Singh(2021)}]{kawintiranon-singh-2021-knowledge}
Kawintiranon, K.; and Singh, L. 2021.
\newblock Knowledge Enhanced Masked Language Model for Stance Detection.
\newblock In Toutanova, K.; Rumshisky, A.; Zettlemoyer, L.; Hakkani-Tur, D.; Beltagy, I.; Bethard, S.; Cotterell, R.; Chakraborty, T.; and Zhou, Y., eds., \emph{Proceedings of the 2021 Conference of the North American Chapter of the Association for Computational Linguistics: Human Language Technologies}, 4725--4735. Online: Association for Computational Linguistics.

\bibitem[{Khandelwal(2021)}]{Khandelwal_2021}
Khandelwal, A. 2021.
\newblock Fine-Tune Longformer for Jointly Predicting Rumor Stance and Veracity.
\newblock In \emph{Proceedings of the 3rd ACM India Joint International Conference on Data Science \& Management of Data (8th ACM IKDD CODS \& 26th COMAD)}, CODS-COMAD ’21, 10–19. New York, NY, USA: Association for Computing Machinery.
\newblock ISBN 978-1-4503-8817-7.

\bibitem[{Küçük and Can(2020)}]{Küçük_Can_2020}
Küçük, D.; and Can, F. 2020.
\newblock Stance Detection: A Survey.
\newblock \emph{ACM Computing Surveys}, 53(1): 12:1--12:37.

\bibitem[{Lan et~al.(2024)Lan, Gao, Jin, and Li}]{lan2024stance}
Lan, X.; Gao, C.; Jin, D.; and Li, Y. 2024.
\newblock Stance detection with collaborative role-infused llm-based agents.
\newblock In \emph{Proceedings of the International AAAI Conference on Web and Social Media}, volume~18, 891--903.

\bibitem[{Li and Caragea(2019)}]{li2019multi}
Li, Y.; and Caragea, C. 2019.
\newblock Multi-task stance detection with sentiment and stance lexicons.
\newblock In \emph{Proceedings of the 2019 conference on empirical methods in natural language processing and the 9th international joint conference on natural language processing (EMNLP-IJCNLP)}, 6299--6305.

\bibitem[{Liang et~al.(2024)Liang, Li, Zhao, Gui, Yang, Yu, Wong, and Xu}]{Liang_et_al_2024}
Liang, B.; Li, A.; Zhao, J.; Gui, L.; Yang, M.; Yu, Y.; Wong, K.-F.; and Xu, R. 2024.
\newblock Multi-modal Stance Detection: New Datasets and Model.
\newblock (arXiv:2402.14298).
\newblock ArXiv:2402.14298.

\bibitem[{Liyanage, Gokani, and Mago(2023)}]{liyanage2023gpt}
Liyanage, C.; Gokani, R.; and Mago, V. 2023.
\newblock GPT-4 as a Twitter data annotator: Unraveling its performance on a stance classification task.
\newblock \emph{Authorea Preprints}.

\bibitem[{Matsa and Eva(2015)}]{Matsa_2015}
Matsa, J.~G., Amy~Mitchell; and Eva, K. 2015.
\newblock Millennials and Political News.

\bibitem[{McNemar(1947)}]{McNemar_1947}
McNemar, Q. 1947.
\newblock Note on the Sampling Error of the Difference Between Correlated Proportions or Percentages.
\newblock \emph{Psychometrika}, 12(2): 153–157.

\bibitem[{Mohammad, Sobhani, and Kiritchenko(2017)}]{mohammad2017stance}
Mohammad, S.~M.; Sobhani, P.; and Kiritchenko, S. 2017.
\newblock Stance and sentiment in tweets.
\newblock \emph{ACM Transactions on Internet Technology (TOIT)}, 17(3): 1--23.

\bibitem[{Niu et~al.(2024)Niu, Cheng, Fu, Peng, Dai, Chen, Huang, and Zhang}]{Niu_et_al_2024}
Niu, F.; Cheng, Z.; Fu, X.; Peng, X.; Dai, G.; Chen, Y.; Huang, H.; and Zhang, B. 2024.
\newblock Multimodal Multi-turn Conversation Stance Detection: A Challenge Dataset and Effective Model.
\newblock In \emph{Proceedings of the 32nd ACM International Conference on Multimedia}, MM ’24, 3867–3876. New York, NY, USA: Association for Computing Machinery.
\newblock ISBN 9798400706868.

\bibitem[{Paszke et~al.(2019)Paszke, Gross, Massa, Lerer, Bradbury, Chanan, Killeen, Lin, Gimelshein, Antiga, Desmaison, Kopf, Yang, DeVito, Raison, Tejani, Chilamkurthy, Steiner, Fang, Bai, and Chintala}]{pytorch}
Paszke, A.; Gross, S.; Massa, F.; Lerer, A.; Bradbury, J.; Chanan, G.; Killeen, T.; Lin, Z.; Gimelshein, N.; Antiga, L.; Desmaison, A.; Kopf, A.; Yang, E.; DeVito, Z.; Raison, M.; Tejani, A.; Chilamkurthy, S.; Steiner, B.; Fang, L.; Bai, J.; and Chintala, S. 2019.
\newblock PyTorch: An Imperative Style, High-Performance Deep Learning Library.
\newblock In \emph{Advances in Neural Information Processing Systems 32}, 8024--8035. Curran Associates, Inc.

\bibitem[{Radford et~al.(2021)Radford, Kim, Hallacy, Ramesh, Goh, Agarwal, Sastry, Askell, Mishkin, Clark, Krueger, and Sutskever}]{Radford_et_al_2021}
Radford, A.; Kim, J.~W.; Hallacy, C.; Ramesh, A.; Goh, G.; Agarwal, S.; Sastry, G.; Askell, A.; Mishkin, P.; Clark, J.; Krueger, G.; and Sutskever, I. 2021.
\newblock Learning Transferable Visual Models From Natural Language Supervision.
\newblock In \emph{Proceedings of the 38th International Conference on Machine Learning}, 8748–8763. PMLR.

\bibitem[{Siddiqua, Chy, and Aono(2019)}]{siddiqua2019tweet}
Siddiqua, U.~A.; Chy, A.~N.; and Aono, M. 2019.
\newblock Tweet stance detection using an attention based neural ensemble model.
\newblock In \emph{Proceedings of the 2019 conference of the north American chapter of the association for computational linguistics: Human language technologies, volume 1 (long and short papers)}, 1868--1873.

\bibitem[{Singh, Bontcheva, and Scarton(2024)}]{Singh_Bontcheva_Scarton_2024}
Singh, I.; Bontcheva, K.; and Scarton, C. 2024.
\newblock The False COVID-19 Narratives That Keep Being Debunked: A Spatiotemporal Analysis.
\newblock (arXiv:2107.12303).
\newblock ArXiv:2107.12303 [cs].

\bibitem[{Suppa et~al.(2024)Suppa, Skala, Jass, Sucik, Svec, and Hraska}]{suppa-etal-2024-bryndza}
Suppa, M.; Skala, D.; Jass, D.; Sucik, S.; Svec, A.; and Hraska, P. 2024.
\newblock Bryndza at {C}limate{A}ctivism 2024: Stance, Target and Hate Event Detection via Retrieval-Augmented {GPT}-4 and {LL}a{MA}.
\newblock In H{\"u}rriyeto{\u{g}}lu, A.; Tanev, H.; Thapa, S.; and Uludo{\u{g}}an, G., eds., \emph{Proceedings of the 7th Workshop on Challenges and Applications of Automated Extraction of Socio-political Events from Text (CASE 2024)}, 166--177. St. Julians, Malta: Association for Computational Linguistics.

\bibitem[{Taira et~al.(2021)Taira, Kreger, Orue, and Diamond}]{Taira_Kreger_Orue_Diamond_2021}
Taira, B.~R.; Kreger, V.; Orue, A.; and Diamond, L.~C. 2021.
\newblock A Pragmatic Assessment of Google Translate for Emergency Department Instructions.
\newblock \emph{Journal of General Internal Medicine}, 36(11): 3361–3365.

\bibitem[{Vamvas and Sennrich(2020)}]{Vamvas_Sennrich_2020}
Vamvas, J.; and Sennrich, R. 2020.
\newblock X-Stance: A Multilingual Multi-Target Dataset for Stance Detection.
\newblock (arXiv:2003.08385).
\newblock ArXiv:2003.08385 [cs].

\bibitem[{Walker et~al.(2012)Walker, Anand, Abbott, Tree, Martell, and King}]{walker2012your}
Walker, M.~A.; Anand, P.; Abbott, R.; Tree, J. E.~F.; Martell, C.; and King, J. 2012.
\newblock That is your evidence?: Classifying stance in online political debate.
\newblock \emph{Decision Support Systems}, 53(4): 719--729.

\bibitem[{Wang et~al.(2024{\natexlab{a}})Wang, Zuo, Peng, and Plank}]{Wang_Zuo_Peng_Plank_2024}
Wang, J.; Zuo, L.; Peng, S.; and Plank, B. 2024{\natexlab{a}}.
\newblock MultiClimate: Multimodal Stance Detection on Climate Change Videos.
\newblock (arXiv:2409.18346).
\newblock ArXiv:2409.18346 [cs].

\bibitem[{Wang et~al.(2024{\natexlab{b}})Wang, Bai, Tan, Wang, Fan, Bai, Chen, Liu, Wang, Ge, Fan, Dang, Du, Ren, Men, Liu, Zhou, Zhou, and Lin}]{Wang_et_al_2024}
Wang, P.; Bai, S.; Tan, S.; Wang, S.; Fan, Z.; Bai, J.; Chen, K.; Liu, X.; Wang, J.; Ge, W.; Fan, Y.; Dang, K.; Du, M.; Ren, X.; Men, R.; Liu, D.; Zhou, C.; Zhou, J.; and Lin, J. 2024{\natexlab{b}}.
\newblock Qwen2-VL: Enhancing Vision-Language Model’s Perception of the World at Any Resolution.
\newblock (arXiv:2409.12191).
\newblock ArXiv:2409.12191 [cs].

\bibitem[{Yin et~al.(2024)Yin, Fu, Zhao, Li, Sun, Xu, and Chen}]{Yin_et_al_2024}
Yin, S.; Fu, C.; Zhao, S.; Li, K.; Sun, X.; Xu, T.; and Chen, E. 2024.
\newblock A Survey on Multimodal Large Language Models.
\newblock \emph{National Science Review}, nwae403.

\bibitem[{Zheng et~al.(2022)Zheng, Baheti, Naous, Xu, and Ritter}]{Zheng_Baheti_Naous_Xu_Ritter_2022}
Zheng, J.; Baheti, A.; Naous, T.; Xu, W.; and Ritter, A. 2022.
\newblock Stanceosaurus: Classifying Stance Towards Multicultural Misinformation.
\newblock In Goldberg, Y.; Kozareva, Z.; and Zhang, Y., eds., \emph{Proceedings of the 2022 Conference on Empirical Methods in Natural Language Processing}, 2132–2151. Abu Dhabi, United Arab Emirates: Association for Computational Linguistics.

\bibitem[{Zotova et~al.(2020)Zotova, Agerri, Nuñez, and Rigau}]{Zotova_Agerri_Nuñez_Rigau_2020}
Zotova, E.; Agerri, R.; Nuñez, M.; and Rigau, G. 2020.
\newblock Multilingual Stance Detection in Tweets: The Catalonia Independence Corpus.
\newblock In Calzolari, N.; Béchet, F.; Blache, P.; Choukri, K.; Cieri, C.; Declerck, T.; Goggi, S.; Isahara, H.; Maegaard, B.; Mariani, J.; Mazo, H.; Moreno, A.; Odijk, J.; and Piperidis, S., eds., \emph{Proceedings of the Twelfth Language Resources and Evaluation Conference}, 1368–1375. Marseille, France: European Language Resources Association.
\newblock ISBN 979-10-95546-34-4.

\end{thebibliography}

\appendix

\section{Details of the Language Support Evaluations}
\label{app:lang_support_details}

We here provide additional details regarding our evaluation of the languages supported by each VLM in this study, given in \cref{tab:language_support}.

\subsection{Evaluation Tasks}

The goal of this evaluation was to gain a preliminary understanding of how well each VLM understands each of the seven target languages. We emphasize that this evaluation was meant only to provide context to our findings regarding multilinguality, and that a more in-depth analysis would be required to draw confident conclusions regarding language support in each model.

The evaluation consisted of 5 tasks, each comprised of a single example. A given model was deemed to \textit{pass} a given task if its output was 1) fluent in the language expected by the task 2) demonstrated an understanding of the task. E.g., output that is fluent but unrelated to the input resulted in a \textit{fail}. The tasks are given below in their English version. Translations were obtained using Google Translate and their consistency with the English was validated using back-translation to English.

\paragraph{Text \& Image Tasks:}
\begin{enumerate}
    \item {Describe the image.}
    \item {What does this animal eat?}
\end{enumerate}

\paragraph{Text Only Tasks:}
\begin{enumerate}\addtocounter{enumi}{2}
    \item {What does a cat eat?}
    \item {Translate the following sentence into \{target language\}: This is my favorite kind of cat, with sleek fur and large ears.}\label{task:2}
    \item {Translate the following sentence into English: \{sentence from \cref{task:2} in target language\}}
\end{enumerate}

\noindent
The Text \& Image tasks were provided the prompt shown as well as a picture of a cat. The Text Only tasks were provided with the prompt only.

As a control, we evaluated two additional low-resource languages that we were certain the VLMs do not know: Udmurt (an Uralic language spoken in Udmurtia, a small republic within Russia) and Yucatec Maya (a Mayan language spoken in the Yucatán Peninsula). For each task, these provided an example of model output for an unknown language to which output for the target languages could be compared.

\subsection{Results}

The results of this evaluation for each task and language are given in \cref{tab:lang_support_detail}, which were summarized in \cref{tab:language_support}. We note that while not shown, the output for all models and tasks on Udmurt and Yucatec Maya resulted in a fail.

\begin{table}[ht]
    \centering
    \small
    \begin{tabular}{lc|cccc}
        \toprule
        \mc{2}{c|}{Task}&  intern  & qwen & ovis & llama \\
        \midrule
        en    &   1   &    \ck   &  \ck & \ck  & \ck   \\
              &   2   &    \ck   &  \ck & \ck  & \ck   \\
              &   3   &    \ck   &  \ck & \ck  & \ck   \\
              &   4   &    \ck   &  \ck & \ck  & \ck   \\
              &   5   &    \ck   &  \ck & \ck  & \ck   \\
        \midrule
        de    &   1   &    \ck   &  \ck &  \ck & \ck   \\
              &   2   &    \ck   &  \ck &      & \ck   \\
              &   3   &          &  \ck &  \ck & \ck   \\
              &   4   &          &  \ck &  \ck & \ck   \\
              &   5   &          &  \ck &  \ck & \ck   \\
        \midrule
        es    &   1   &    \ck   &  \ck &      & \ck   \\
              &   2   &    \ck   &  \ck &  \ck & \ck   \\
              &   3   &    \ck   &  \ck &  \ck & \ck   \\
              &   4   &          &  \ck &  \ck & \ck   \\
              &   5   &          &  \ck &  \ck & \ck   \\
        \midrule
        fr    &   1   &    \ck   &  \ck &  \ck & \ck   \\
              &   2   &    \ck   &  \ck &      & \ck   \\
              &   3   &          &  \ck &  \ck & \ck   \\
              &   4   &          &  \ck &  \ck & \ck   \\
              &   5   &          &  \ck &  \ck & \ck   \\
        \midrule
        hi    &   1   &          &  \ck &  \ck & \ck   \\
              &   2   &          &      &  \ck & \ck   \\
              &   3   &          &  \ck &  \ck & \ck   \\
              &   4   &          &  \ck &  \ck & \ck   \\
              &   5   &          &      &  \ck & \ck   \\
        \midrule
        pt    &   1   &    \ck   &  \ck &  \ck & \ck   \\
              &   2   &    \ck   &  \ck &  \ck & \ck   \\
              &   3   &    \ck   &  \ck &  \ck & \ck   \\
              &   4   &          &  \ck &  \ck & \ck   \\
              &   5   &          &  \ck &  \ck & \ck   \\
        \midrule
        zh    &   1   &    \ck   &  \ck &  \ck &       \\
              &   2   &    \ck   &  \ck &  \ck &       \\
              &   3   &    \ck   &  \ck &  \ck & \ck   \\
              &   4   &    \ck   &  \ck &  \ck & \ck   \\
              &   5   &    \ck   &  \ck &  \ck & \ck   \\
        \bottomrule
    \end{tabular}
    \caption{Results of prompting each model on each of the language support evaluation tasks. A checkmark \ck indicates a pass. An empty cell indicates a fail.}
    \label{tab:lang_support_detail}
\end{table}

\end{document}